\documentclass{article} 
\usepackage{iclr2026_conference,times}

\usepackage{amsmath,amsfonts,bm}

\def\eqref#1{equation~\ref{#1}}

\def\1{\bm{1}}

\DeclareMathAlphabet{\mathsfit}{\encodingdefault}{\sfdefault}{m}{sl}
\SetMathAlphabet{\mathsfit}{bold}{\encodingdefault}{\sfdefault}{bx}{n}

\usepackage[table]{xcolor}  
\usepackage{hyperref}
\usepackage{url}
\usepackage{multirow}
\usepackage{graphicx}
\usepackage{bm}
\usepackage{amssymb}
\usepackage{booktabs}     
\usepackage{tabularray}
\definecolor{lightblue}{RGB}{217,235,255}

\usepackage{enumitem}
\newcommand\blfootnote[1]{%
  \begingroup
  \hypersetup{hidelinks}%
  \renewcommand\thefootnote{}\footnote{#1}%
  \addtocounter{footnote}{-1}%
  \endgroup
}
\hypersetup{
    colorlinks=false,
    pdfborder={0 0 0}
}

\title{UniSplat: Unified Spatio-Temporal Fusion via 3D Latent Scaffolds for Dynamic Driving Scene Reconstruction}

\author{
Chen Shi$^{1,2}$ \quad Shaoshuai Shi$^2$ \quad Xiaoyang Lyu$^3$ \\
Chunyang Liu$^2$ \quad Kehua Sheng$^2$ \quad Bo Zhang$^2$ \quad Li Jiang$^{1\dagger}$ \\[1ex]
{\small $^1$The Chinese University of Hong Kong, Shenzhen} \\
{\small $^2$Voyager Research, Didi Chuxing \quad $^3$The University of Hong Kong} \\[3ex]
\textbf{Project Page:} \href{https://chenshi3.github.io/unisplat.github.io/}{\small\texttt{https://chenshi3.github.io/unisplat.github.io/}}
}

\iclrfinalcopy 
\begin{document}

\maketitle

\begin{abstract}

Feed-forward 3D reconstruction for autonomous driving has advanced rapidly, yet existing methods struggle with the joint challenges of sparse, non-overlapping camera views and complex scene dynamics. 
We present UniSplat, a general feed-forward framework that learns robust dynamic scene reconstruction through unified latent spatio-temporal fusion. UniSplat constructs a 3D latent scaffold, a structured representation that captures geometric and semantic scene context by leveraging pretrained foundation models. To effectively integrate information across spatial views and temporal frames, we introduce an efficient fusion mechanism that operates directly within the 3D scaffold, enabling consistent spatio-temporal alignment. To ensure complete and detailed reconstructions, we design a dual-branch decoder that generates dynamic-aware Gaussians from the fused scaffold by combining point-anchored refinement with voxel-based generation, and maintain a persistent memory of static Gaussians to enable streaming scene completion beyond current camera coverage. Extensive experiments on real-world datasets demonstrate that UniSplat achieves state-of-the-art performance in novel view synthesis, while providing robust and high-quality renderings even for viewpoints outside the original camera coverage.

\end{abstract}
\blfootnote{$\dagger$: Corresponding author.}
\section{INTRODUCTION}

Replicating 3D scenes from urban driving sequences has emerged as a core capability for autonomous systems, supporting simulation~\citep{Cao2025CORL,yang2023unisim,tonderski2024neurad}, scene understanding~\citep{huang2024gaussianformer,huang2025gaussianformer,yan2025st}, and long-horizon planning~\citep {murai2025mast3r}. Recent advances in 3D Gaussian Splatting \citep{yan2024street,xu2025adgs,kerbl20233d} 
have demonstrated impressive rendering efficiency and fidelity. 
However, these methods typically assume substantial viewpoint overlap among input images and rely on per-scene optimization, 
which limits their applicability in real-time driving scenarios.

To enable faster inference, feed-forward reconstruction methods have emerged to synthesize novel views in a single forward pass~\citep{xu2025depthsplat,chen2024mvsplat,zhang2025transplat,lu2024drivingrecon}. These methods typically encode inter-view correlations within the image domain via cross-attention or by constructing a multi-view stereo (MVS) cost volume, and subsequently decode the Gaussian primitives from the resulting fused features. Notably, the choice of fusion strategy is crucial, as it significantly impacts the final rendering quality. EvolSplat~\citep{miao2025evolsplat} integrates multi-frame geometric information from front-view monocular sequences using 3D-CNN, but ignores semantic fusion and lacks mechanisms for dynamic handling. Meanwhile, Omni-Scene~\citep{wei2025omni} leverages a Triplane Transformer for strong multi-view fusion but does not incorporate temporal aggregation and is constrained by coarse-grained 3D details. 
Despite these advances, robust reconstruction in urban driving scenarios remains challenging, particularly in maintaining a unified latent representation that evolves smoothly over time, handling partial observations, occlusions, and dynamic motion, and efficiently generating high-fidelity Gaussians from sparse inputs.

To address these challenges, we propose \textbf{UniSplat}, a general feed-forward framework for dynamic scene modeling from multi-camera videos. The core insight of UniSplat is to construct a unified 3D scaffold that fuses both multi-view spatial information and multi-frame temporal information. This scaffold facilitates geometric and semantic contextual interaction in 3D space, supports efficient long-term information integration and dynamic modeling, and enables effective decoding of Gaussian primitives. 
By preserving and fusing essential information, it ensures coherent and consistent scene reconstruction over time.

Specifically, the UniSplat framework follows a three-stage pipeline. First, we construct an ego-centric 3D scaffold by feeding multi-view images to a pretrained geometry foundation model and a visual foundation model, encoding both geometry structure and semantics cues into a sprase 3D feature volume. Second, we perform spatio-temporal fusion by integrating multi-view spatial context within the current frame's scaffolds and fusing historical scaffolds into current scaffolds via ego-motion compensation, yielding a temporal-enhanced scene representation.
Third, we decode the fused scaffold into Gaussians via a dual-branch strategy: 
one branch predicts Gaussians at sparse point locations for fine-grained detail
while the other directly generates new Gaussians from voxel centers to complement point anchor predictions. Each Gaussian is assigned a dynamic probability score to identify static content, allowing us to maintain a memory bank of persistent static Gaussians across frames for long-term scene completion.

We evaluate our method on the Waymo Open dataset~\citep{sun2020scalability} and NuScenes~\citep{caesar2020nuscenes} dataset, which present dynamic street scenes with complex environmental conditions and limited overlap for multi-camera images. Experimental results demonstrate that our approach achieves state-of-the-art performance across both datasets in input-view reconstruction and novel-view synthesis. Notably, with the help of temporal memory, our model exhibits strong robustness and superior rendering quality when synthesizing views outside the original camera coverage. 

In summary, our main contributions are as follows:
\begin{itemize}
\item We introduce UniSplat, a novel feed-forward framework for dynamic scene reconstruction from multi-camera videos via a unified 3D latent scaffold.
\item We design a novel scaffold-based fusion mechanism that supports unified spatio-temporal alignment and progressive scene memory integration.
\item We propose a dual-branch Gaussian generation mechanism with dynamic-aware filtering, enabling fine-grained and complete rendering and memory-based scene completion.
\item Comprehensive experiments on two large-scale driving datasets demonstrate that UniSplat significantly outperforms state-of-the-art feed-forward reconstruction methods, with generalization capability for challenging views outside the observed camera frustums.
\end{itemize}

\section{RELATED WORK}
\textbf{Neural 3D Reconstruction.} The field of neural 3D reconstruction has witnessed remarkable progress, largely driven by Neural Radiance Fields (NeRF)~\citep{mildenhall2021nerf} and, more recently, 3D Gaussian Splatting (3DGS)~\citep{kerbl20233d}. NeRF represents scenes as continuous volumetric functions, achieving high-fidelity renderings but incurring substantial computational costs. Subsequently, 3DGS introduced explicit point-based representations with highly efficient rasterization, enabling real-time rendering.  Despite the impressive performance of NeRF, 3DGS, and their extensive variants~\citep{hu2023tri, xu2022point, muller2022instant,hu2023tri,yu2024mip,yang2025improving}, these methods are usually limited by the reliance on dense input views and costly per-scene optimization, thereby restricting their scalability. Alternatively, feed-forward methods tackle this challenge by learning generalizable scene priors from large-scale datasets during training, allowing for immediate reconstruction from sparse observations at inference time. MuRF~\citep{xu2024murf} employs target view frustum volumes for radiance field reconstruction. PixelSplat~\citep{charatan2024pixelsplat} and Splatter Image~\citep{szymanowicz2024splatter} predict per-pixel 3D Gaussians from image features, while MVSplat~\citep{chen2024mvsplat} leverages cost volumes for geometric consistency and DepthSplat~\citep{xu2025depthsplat} integrates features from pre-trained monocular depth models to improve robustness. However, these approaches still face significant challenges in complex urban driving scenarios. The minimal overlap among surround-view cameras compromises multi-view correspondence, while the presence of highly dynamic objects complicates temporal aggregation. In this work, we develop a feed-forward framework to reconstruct complete driving scenes from sparse views while effectively leveraging multi-frame information.

\textbf{Driving Scene Reconstruction with 3D Gaussians.}  Leveraging advances in 3D Gaussian Splatting, several works~\citep{ chen2023periodic,huang2024s3gaussian,zhou2024drivinggaussian,yan2024street,zhao2025drivedreamer4d,yan2025streetcrafter,fan2025freesim,xu2025adgs} specialize in driving scenes, focusing on 3D or 4D reconstruction within individual scenes through offline optimization. In parallel, generalizable methods have also emerged. These approaches~\cite{tian2025drivingforward,lu2024drivingrecon} typically employ depth networks to determine Gaussian primitive positions in a feed-forward manner and predict per-pixel Gaussians along camera rays. To enhance global consistency and completeness, several techniques further incorporate 3D spatial representations. EVolSplat~\citep{miao2025evolsplat} directly accumulates depth across multiple frames and leverages 3D-CNNs to refine Gaussian geometry. Omni-Scene~\citep{wei2025omni} transforms multi-view image features into TriPlane representations and decodes voxel-anchored Gaussians to complement pixel-based estimates. SCube~\citep{ren2024scube} constructs a detailed sparse-voxel scaffold via a hierarchical voxel latent diffusion model. Despite the progress made, these methods often focus on static or single-frame reconstruction and  struggle to simultaneously handle multi-view fusion and dynamic multi-frame aggregation. To counter these challenges, we propose UniSplat, a novel framework that unifies multi-view fusion and dynamic multi-frame aggregation within a 3D latent scaffold.

\textbf{3D Geometry Reconstruction.} End-to-end, data-driven pipelines that reconstruct scene geometry directly from images have progressed rapidly. DUSt3R~\citep{wang2024dust3r} pioneers a transformer-based framework that predicts 3D point maps from uncalibrated image pairs. Subsequent works~\citep{wang2025pi,wang2025vggt,yang2025fast3r,wang2025continuous,chen2025long3r,xiao2025spatialtrackerv2} extend this paradigm to arbitrary multi-view inputs and scale up both training data and model capacity, achieving state-of-the-art reconstruction accuracy with strong generalization across diverse scenes. However, these methods generally struggle with poor texture representation and encounter multi-view misalignment under minimal overlap, limiting novel view rendering quality. In this work, we employ these 3D foundation models to obtain a geometry initialization from images, and then perform 3D alignment and fusion in the learned latent scaffold.

\section{UniSplat}

\begin{figure}[h]
\begin{center}
\includegraphics[width=0.99\textwidth]{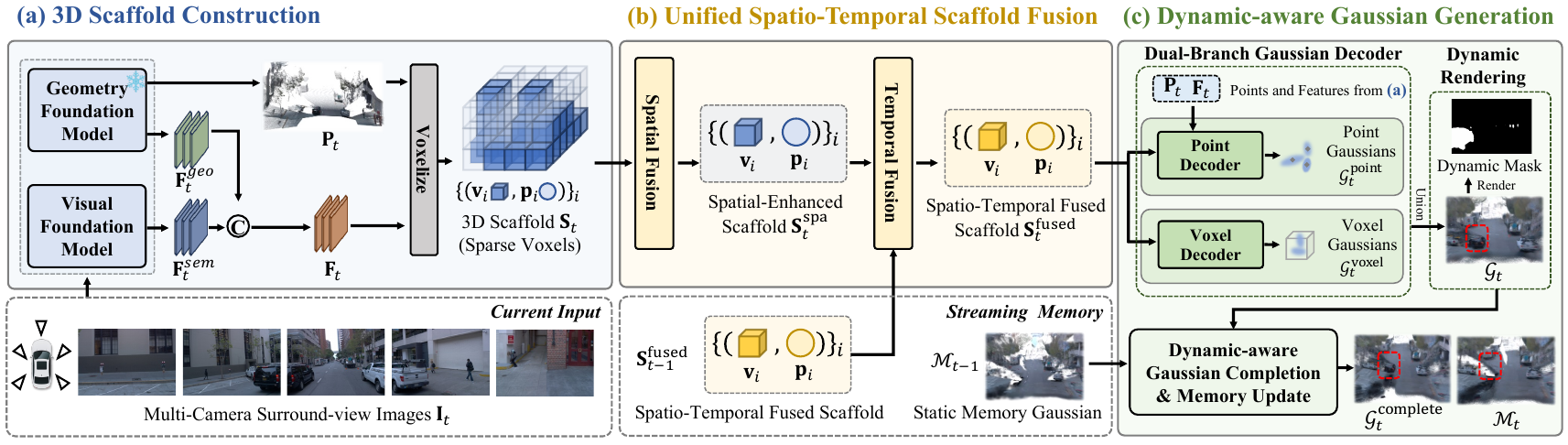}
\end{center}
\vspace{-15pt}
\caption{\textbf{Overview of UniSplat.} Given multi-camera images from vehicle-mounted cameras, UniSplat leverages foundation models to construct geometry-semantic aware 3D latent scaffolds, where unified spatio-temporal fusion is performed. From this scaffold, a dual-branch decoder generates dynamic-aware Gaussian primitives using both point anchors and voxel centers, with dynamic filtering maintaining a persistent memory of static scene content. The \textcolor{red}{red boxes} highlight a dynamic car that is filtered out in our memory module (best viewed when zoomed in).}
\label{fig:pipeline}
\end{figure}

UniSplat operates on a continuous stream of multi-camera frames, maintaining a unified 3D latent representation of the scene that evolves over time. As shown in Fig.~\ref{fig:pipeline}, each time step begins with 3D scaffold construction from multi-view images (Sec.~\ref{method-part1}), producing a set of 3D voxels (the latent scaffold) that encodes the scene’s geometry and semantics in an ego-centric coordinate frame. We then perform a unified spatio-temporal fusion, integrating information across views within the current scaffold and aggregating it with the latent scaffold from the previous time step (Sec.~\ref{method-part2}). Finally, we achieve dynamic-aware Gaussian generation (Sec.~\ref{method-part3}) through a dual-branch decoder that estimates dynamic-aware Gaussian primitives from both points and voxels, while maintaining a temporal memory bank that accumulates static Gaussians over time to address incomplete scene coverage caused by sparse camera inputs and limited fields of view.

\subsection{Preliminary} \label{prelim}
3D Gaussian Splatting~\citep{kerbl20233d} represents a scene as a collection of 3D Gaussian primitives $\mathcal{G} = \{G_i\}_{i=1}^{N}$. Each primitive $G_i$ is defined by a tuple of learnable parameters $\theta_i = \{\boldsymbol{\mu}_i, \alpha_i, \mathbf{\Sigma}_i, \mathbf{c}_i\}$, representing its 3D center position, opacity, covariance matrix, and color coefficients, respectively. To render an image from a target viewpoint, these 3D Gaussians are projected onto the 2D image plane and blended using differentiable alpha compositing. Specifically, for a particular pixel, the color contribution $C$ from all Gaussians whose projections cover that pixel is:
\begin{align}
    C = \sum_{i \in \mathcal{N}} \mathbf{c}_i \alpha_i \prod_{j=1}^{i-1} (1 - \alpha_j),
    \label{eq:rendering}
\end{align}
where $\mathcal{N}$ is the set of Gaussians overlapping the pixel, sorted by depth. Beyond simple color rendering, several works~\citep{zhou2024feature,zuo2025fmgs} augment Gaussians with additional parameters,
which can be rendered into a 2D feature map using the same alpha compositing mechanism, enabling the distillation of knowledge from 2D foundation models. Inspired by this extensibility, we introduce a learnable dynamic attribute for each Gaussian to explicitly disentangle scene dynamics.

\subsection{3D Scaffold Construction} \label{method-part1}
Constructing an accurate 3D scaffold from sparse, minimally overlapping camera views is a primary challenge in multi-view reconstruction for driving scenes. To address this, we harness the power of geometry foundation models to infer a coherent 3D structure from multi-view images in one forward pass. We then enrich this 3D geometric scaffold with semantic information from a visual foundation model. This process yields a latent scaffold representation in the ego-centric coordinate frame of the vehicle, which provides a strong basis for subsequent spatio-temporal fusion.

\textbf{Metric-Scale 3D Geometry Generation.} 
%
Given synchronized multi-view images $\mathbf{I}_t = \{I_t^k\}_{k=1}^{N_{\text{cam}}}$ from a multi-camera rig, we apply a state-of-the-art feed-forward multi-view geometry foundation models (\textit{e.g.}, \citet{wang2025vggt, wang2025pi}) to directly predict a dense 3D point map $\mathbf{P}_t^{\text{init}} \in \mathbb{R}^{N_{\text{cam}} \times H_{\text{img}} \times W_{\text{img}} \times 3}$, where each pixel is associated with a 3D coordinate inferred jointly from all views. Unlike per-view depth estimation and late fusion, this unified approach leverages learned multi-view correspondences to generate a coherent scene-level point cloud. 
However, such predictions often suffer from scale ambiguity, which is problematic in autonomous driving. Thus, we introduce an auxiliary scale alignment branch: a small MLP predicts per-camera scale factors from the pooled geometry features:
\begin{align}
    \boldsymbol{\gamma} = \operatorname{MLP}(\operatorname{AvgPool}(\mathbf{F}_t^{\text{geo}}, \{H, W\})) \in \mathbb{R}^{N_{\text{cam}}}, 
    \label{eq:scale_prediction}
\end{align}
where $\mathbf{F}_t^{\text{geo}}$ denotes the hidden feature maps from the geometry model and $\operatorname{AvgPool}(\cdot, \{H, W\})$ represents the average pooling over the height and width dimensions for each view. The scale prediction is supervised by minimizing the error between $\boldsymbol{\gamma}$ and the optimal scale vector computed using the ROE solver~\citep{wang2025mogev1} with LiDAR point references. Applying $\boldsymbol{\gamma}$ to $\mathbf{P}_t^{\text{init}}$ yields a metric-consistent point cloud $\mathbf{P}_t$ as the geometric foundation of our scaffold.

\textbf{Scaffold Construction with Geometric-Semantic Context.} 
As the generated $\mathbf{P}_t$ is an unstructured point set, we organize these points into a sparse voxel grid and fuse geometric and semantic features to create the 3D latent scaffold. To achieve this, we first extract semantic-aware 2D features $\mathbf{F}_t^{\text{sem}}$ from the input views using a visual foundation model~\citep{oquab2023dinov2}, and fuse them with geometric features $\mathbf{F}_t^{\text{geo}}$ to obtain a unified multi-view feature map $\mathbf{F}_t$. We then voxelize the point cloud into $N_{v}$ voxels within 
an ego-centric cuboid $[\mathbf{p}_{\min} \in \mathbb{R}^3, \mathbf{p}_{\max} \in \mathbb{R}^3]$ covering the surrounding scene. The volume is partitioned into voxels of size $\boldsymbol{\epsilon}$, and only voxels containing points are considered valid.
Specifically, for each voxel $i$, we compute its coarse geometric voxel feature $\mathbf{v}_i^{\text{init}}$ as the average of the coordinates of points $j \in \mathcal{I}_i$ that lie in that voxel:
\begin{align}
    \mathbf{v}_i^{\text{init}} =  \frac{\sum_{j \in \mathcal{I}_i}  \mathbf{P}_{t,j}}{\sum_{j \in \mathcal{I}_i} 1}, \quad i \in \{1, \ldots, N_{v}\}, 
    \label{eq:voxel_aggregation}
\end{align}
where $\mathcal{I}_i$ is the index set of points within $i\text{-th}$ voxel. 
Next, to enrich the voxel with geometric-semantic context, we project each voxel center into the input views and sample the corresponding features from $\mathbf{F}_t$, which are then concatenated with the initial voxel feature $\mathbf{v}_i^{\text{init}}$.
The resulting 3D scaffold $\mathbf{S}_t$ of the scene is formally defined as a set of these voxels:
\begin{align}
    \mathbf{S}_t = \{(\mathbf{v}_i\in\mathbb{R}^{C_s}, \mathbf{p}_i\in\mathbb{R}^{3})\}_{i=1}^{N_{v}}
\end{align}
where $C_s$ is the feature dimension, $\mathbf{v}_i$ represents the voxel feature encoding both geometric and semantic context, and $\mathbf{p}_i$ denotes the corresponding voxel center that preserves explicit 3D structure.


\subsection{Unified Spatio-Temporal Scaffold Fusion} \label{method-part2}
A key advantage of our scaffold representation lies in its inherent structure, which encodes explicit 3D geometry within a unified ego-centric space. This design enables contextual interaction in the unified 3D space, supporting direct and efficient spatio-temporal fusion across multiple views and temporal frames within a single scaffold representation.

\textbf{Spatial Scaffold Fusion.} Unlike traditional approaches~\citet{chen2024mvsplat,xu2025depthsplat} that fuse spatial information across views in 2D space using image-level cross-attention, which is often hindered by limited overlap between views, we perform spatial fusion directly in the 3D scaffold space. In this representation, spatially corresponding information from different views is naturally aligned in 3D space. Specifically, we employ a sparse 3D U-Net $\phi$ to integrate multi-view features, producing a spatially-enhanced scaffold representation $\mathbf{S}_{t}^\text{spa}$:
\begin{align}
\quad \mathbf{S}_{t}^\text{spa} = \phi(\mathbf{S}_t),
\end{align}

\textbf{Temporal Scaffold Fusion.} Instead of processing historical raw images as in existing works~\citep{lu2024drivingrecon, tian2025drivingforward}, we integrate temporal cues directly within the scaffold representation in a streaming manner. 
Given the previous fused latent scaffold features $\mathbf{S}_{t-1}^\text{fused}$ from a streaming memory, we first warp its voxel centers into the current frame’s coordinate system using the known ego-pose $T_{t-1}^t$, and their features are tagged with a time-step embedding to distinguish them from current observations. We then merge the transformed previous scaffold $\mathbf{S}_{t-1}^\text{fused}$ with the current scaffold $\mathbf{S}_{t}^\text{spa}$ via element-wise addition at any overlapping voxels, and simply union the features for non-overlapping regions. We denote this operation as a sparse tensor addition:
\begin{align}
 \mathbf{S}_{t}^\text{fused} = \mathbf{S}_{t}^\text{spa} \oplus \operatorname{Warp}(\mathbf{S}_{t-1}^\text{fused},T_{t-1}^{t})
\end{align}
where $\oplus$ denotes sparse tensor addition that aggregates features at overlapping voxel locations while preserving non-overlapping features from both sparse tensors.  The resulting tensor $\mathbf{S}_{t}^\text{fused}$ is further refined by a lightweight sparse convolutional network to capture complex temporal dependencies and is cached back into the streaming memory to maintain long-term temporal information. 


\subsection{Dynamic-aware Gaussian Generation}  \label{method-part3}
Building upon the spatio-temporally fused scaffold $\mathbf{S}_{t}^\text{fused}$, we generate a set of 3D Gaussian primitives via a dual-branch decoding strategy, yielding primitives that explicitly disentangle static and dynamic scene components, which enables progressive scene completion over time.

\textbf{Dual-Branch Gaussian Decoder.} 
Our Gaussian decoder comprises two complementary branches that jointly enhance reconstruction fidelity and completeness.
The point decoder branch focuses on preserving fine-grained geometric details by leveraging the point-level anchors from the reconstructed metric-scale point map $\mathbf{P}_t$. For each point $\mathbf{P}_{t,i} \in \mathbf{P}_t$, we locate its voxel coordinate in the scaffold and retrieve the corresponding latent feature from $\mathbf{S}_{t}^\text{fused}$ as:
\begin{align}
 f_{t, i}^{\text{3d}} = \operatorname{Retrieve}\left(\mathbf{S}_{t}^\text{fused}, \left\lfloor \frac{\mathbf{P}_{t,i} - \mathbf{p}_{\min}}{\mathbf{\epsilon}} \right\rfloor\right),
\end{align}
where $\lfloor \cdot \rfloor$ denotes the voxel indexing operation. If a point falls outside the scaffold's spatial extent, zero-padding is applied. Since each point $\mathbf{P}_{t,i}$ maintains a one-to-one correspondence with its source pixel, we additionally sample 2D image feature $f_{t, i}^{\text{2d}}$ for each point from the multi-view feature maps $\mathbf{F}_t$. These features are concatenated to predict the Gaussian primitives via an MLP:
\begin{align}
\{(\Delta\boldsymbol{\mu}_{i}, \alpha_{i}, \mathbf{\Sigma}_i, \mathbf{c}_{i}, d_{i})\} = \operatorname{MLP}([f_{t, i}^{\text{3d}}, f_{t, i}^{\text{2d}}]),
\label{eq:gs-decode}
\end{align}
where $\Delta\boldsymbol{\mu}_i$ denotes the Gaussian's offset from the point anchor, and $d_i \in \mathbb{R}$ is a learned dynamic score indicating motion likelihood. This branch yields a detailed set of Gaussians denoted as $\mathcal{G}_t^{\text{point}}$.

The voxel decoder branch complements the point-based decoding by directly predicting new Gaussian primitives from voxel-level scaffold features, effectively filling in sparsely covered regions and enhancing the scene completeness.
For each voxel in $\mathbf{S}_{t}^\text{fused}$, we adopt a compact MLP to produce $g$ sets of Gaussian parameters (as in Eq.~\ref{eq:gs-decode}) per voxel. 
The center of each Gaussian is derived by adding the predicted displacement to the voxel center, forming the set $\mathcal{G}_t^{\text{voxel}}$. 
The complete reconstruction at time $t$ is then given by $\mathcal{G}_t = \mathcal{G}_t^{\text{point}} \cup \mathcal{G}_t^{\text{voxel}}$.

%
\textbf{Dynamic-aware Gaussian Completion.}
To enhance temporal consistency and alleviate occlusion-induced sparsity, we introduce a memory mechanism that maintains accumulated static Gaussians over time. Each Gaussian primitive is associated with a dynamic attribute $d_i$, enabling motion-aware filtering.
Given the static memory $\mathcal{M}_{t-1}$ from the previous frame, we transform it into the current ego-centric coordinate system and perform a view filtering to remove Gaussians visible in the current field of view. The resulting filtered memory $\mathcal{M}_{t-1}^{\prime}$ is then fused with the current reconstruction:
\begin{align}
   \mathcal{G}_t^{\text{complete}} = \mathcal{G}_t \cup \mathcal{M}_{t-1}^{\prime} 
\end{align}
where $\mathcal{G}_t^{\text{complete}}$ provides a comprehensive scene representation that fills in the blind spots of the current frame's reconstruction. 
Finally, the memory is updated by retaining static Gaussians from the current frame:
\begin{align}
\mathcal{M}_t = \mathcal{M}_{t-1}^{\prime} \cup \{G_i \in \mathcal{G}_t \mid d_i < \tau_d\}, ~~ i \in \{1, ..., N_{\mathcal{G}_t}\}
\end{align}
where $\tau_d$ is a score threshold, and $N_{\mathcal{G}_t}$ is the total number of current Gaussians. This streaming mechanism enables temporally persistent reconstruction while suppressing dynamic artifacts.

\subsection{Training Objective}
The model is optimized via a composite loss function defined over the rendered outputs from $\mathcal{G}_t$:
\begin{align}
    \mathcal{L} = \sum_{v \in \mathcal{V}_\text{input}} \left( \lambda_1\mathcal{L}_\text{mse}^{v} + \lambda_2 \mathcal{L}_\text{lpips}^{v} + \lambda_3 \mathcal{L}_\text{dyn}^{v} + \lambda_4 \mathcal{L}_\text{scale}^{v} \right) + \sum_{v \in \mathcal{V}_\text{novel}} \left( \lambda_1\mathcal{L}_\text{mse}^{v}  \odot B^v +  \lambda_2 \mathcal{L}_\text{lpips}^{v} \right)
\end{align}
where $\mathcal{L}_\text{mse}^{v}$ and $\mathcal{L}_\text{lpips}^{v}$ are the MSE reconstruction and LPIPS perceptual losses~\citep{zhang2018unreasonable} between rendered and ground-truth images for view $v$, $\mathcal{L}_\text{dyn}^{v}$ is the cross-entropy loss between rendered dynamic scores and ground-truth dynamic segmentation masks, and $\mathcal{L}_\text{scale}^{v}$ is a smooth-L1 loss for scale supervision. $\mathcal{V}_\text{input}$ refers to the set of input camera views at time $t$ and $\mathcal{V}_\text{novel}$ denotes novel viewpoints at time $t+1$. The operator $\odot$ denotes element-wise multiplication, where the background mask $B^v$ excludes dynamic regions to prevent optimization instability.

\section{EXPERIMENTS}
\begin{table}
\caption{Quantitative results on the Waymo Dataset. The best results are marked in \textbf{bold} and \underline{underlined} entries indicate second-place performance. $\text{}^{*}$: Evaluation conducted on front 3 views only. $\dagger$: Results obtained using optimal scale alignment.}
\label{tab:waymo}
\centering
\small 
\setlength{\tabcolsep}{5pt}
\begin{tabular}{l|c|ccc|ccc}
\hline
\multirow{2}{*}{Method} & \multirow{2}{*}{Views} & \multicolumn{3}{c|}{Reconstruction} & \multicolumn{3}{c}{Novel View Synthesis} \\
 & & PSNR$\uparrow$ & SSIM$\uparrow$ & LPIPS$\downarrow$ & PSNR$\uparrow$ & SSIM$\uparrow$ & LPIPS$\downarrow$ \\
\hline
EvolSplat~\citep{miao2025evolsplat} & Front& \underline{23.35} & \underline{0.70}& \underline{0.29} & - & - & - \\
UniSplat & Front& \textbf{28.93}& \textbf{0.86} & \textbf{0.18} & \textbf{27.34} & \textbf{0.80} & \textbf{0.20}\\
\hline
$\text{DriveRecon}^{*}$~\citep{lu2024drivingrecon} & Multi& 23.86 & 0.72 & 0.33 & 17.32 & 0.58 &  0.53\\
MVSplat~\citep{chen2024mvsplat} & Multi& 24.94 & \underline{0.80} & \underline{0.23} & 22.04 & 0.68 & 0.34 \\
DepthSplat~\citep{xu2025depthsplat} & Multi& \underline{25.38} & 0.76 & 0.26 & \underline{23.86} & \underline{0.70} & \underline{0.31} \\
UniSplat & Multi& \textbf{28.56} & \textbf{0.83} & \textbf{0.20} & \textbf{25.12} & \textbf{0.74} & \textbf{0.27} \\
\rowcolor{lightblue}\textcolor{black!50}{$\text{UniSplat}{\dagger}$} & \textcolor{black!50}{Multi}& \textcolor{black!50}{29.58} & \textcolor{black!50}{0.86} & \textcolor{black!50}{0.17} &  \textcolor{black!50}{25.98}& \textcolor{black!50}{0.76}& \textcolor{black!50}{0.24}\\
\hline
\end{tabular}
\vspace{-7pt}
\end{table}

\begin{table}
\centering
\small 
\setlength{\tabcolsep}{20pt} 
\caption{Quantitative results on the nuScenes Dataset. We highlight best results in \textbf{bold} and second-place results with \underline{underlines}. $\text{}^{*}$: reported by \citet{wei2025omni}.}
\label{tab:nuc}
\begin{tabular}{l|ccc}
\hline
Method & PSNR$\uparrow$ & SSIM$\uparrow$ & LPIPS$\downarrow$  \\
\hline
$\text{PixelSplat}^{*}$~\citep{charatan2024pixelsplat} & 21.51 & 0.616 &  0.372  \\
$\text{MVSplat}^{*}$~\citep{chen2024mvsplat}  & 21.61 & 0.658& 0.295  \\
Omin-Scene~\citep{wei2025omni} & \underline{24.27} & \underline{0.736}& \textbf{0.237}  \\
UniSplat & \textbf{25.37}& \textbf{0.765} & \underline{0.246}  \\
\hline
\end{tabular}
\vspace{-10pt}
\end{table}

\subsection{Experimental Settings} \label{Experimental-Settings}
\textbf{Datasets and Metrics.} We conduct experiments on two large-scale autonomous driving benchmarks: Waymo Open~\citep{sun2020scalability} and nuScenes~\citep{caesar2020nuscenes} datasets. The Waymo Open dataset includes 798 training and 202 validation sequences, with all sequences approximately 20 seconds long and captured at 10Hz using five cameras. For nuScenes, which provides six surround-view images per frame, we adopt the strategy of \citet{wei2025omni} and partition scenes into equally spaced bins along the vehicle trajectory, yielding 135,941 training and 30,080 validation bins. Each bin consists of multiple sequential frames, and the central frame serves as the input. 

To measure visual quality, we adopt standard image quality metrics including PSNR, SSIM~\citep{wang2004image}, and LPIPS~\citep{zhang2018unreasonable}. Following \citet{yang2024emernerf,lu2024drivingrecon}, the Waymo benchmark evaluates two tasks: reconstruction, for which images at current timestep $t$ serve as targets, and novel view synthesis, which synthesizes images at the subsequent timestep $t+1$. For nuScenes, consistent with \citet{wei2025omni}, we evaluate on target views consisting of the first, last, and central frames of each bin.

\textbf{Implementation Details.} For our 3D scaffold reconstruction, we employ a frozen pretrained geometry transformer $\pi^3$~\citep{wang2025pi} for initial geometry generation and a pretrained DINOv2 ViT-small backbone~\citep{oquab2023dinov2} for semantic feature extraction. The scaffold is built within a real-world volume of [-72m, -72m, -4m, 72m, 72m, 12m], using an initial voxel size of (0.1m, 0.1m, 0.2m). Scaffold spatial fusion is performed using a sparse 3D U-Net with a maximum downsampling factor of $8\times$, while the temporal fusion employs a separate sparse 3D U-Net with a maximum downsampling factor of $2\times$. In the Gaussian decoding stage, the second branch generates $g = 2$ primitives per voxel, and the dynamic attribute threshold for streaming scene completion is set to $\tau_d = 0.2$. We adopt image resolutions of $350 \times 518$ for the Waymo dataset and $224 \times 406$ for the nuScenes dataset. All models are trained using the AdamW optimizer~\citep{loshchilov2018decoupled} on 16 NVIDIA H20 GPUs with a total batch size of 32. For the training objective,
we set $\lambda_1$=5.0, $\lambda_2$=0.05, $\lambda_3$=0.05, and $\lambda_4$=0.1.

\subsection{Main Results}
\begin{figure}[h]
\begin{center}
\includegraphics[width=0.99\textwidth]{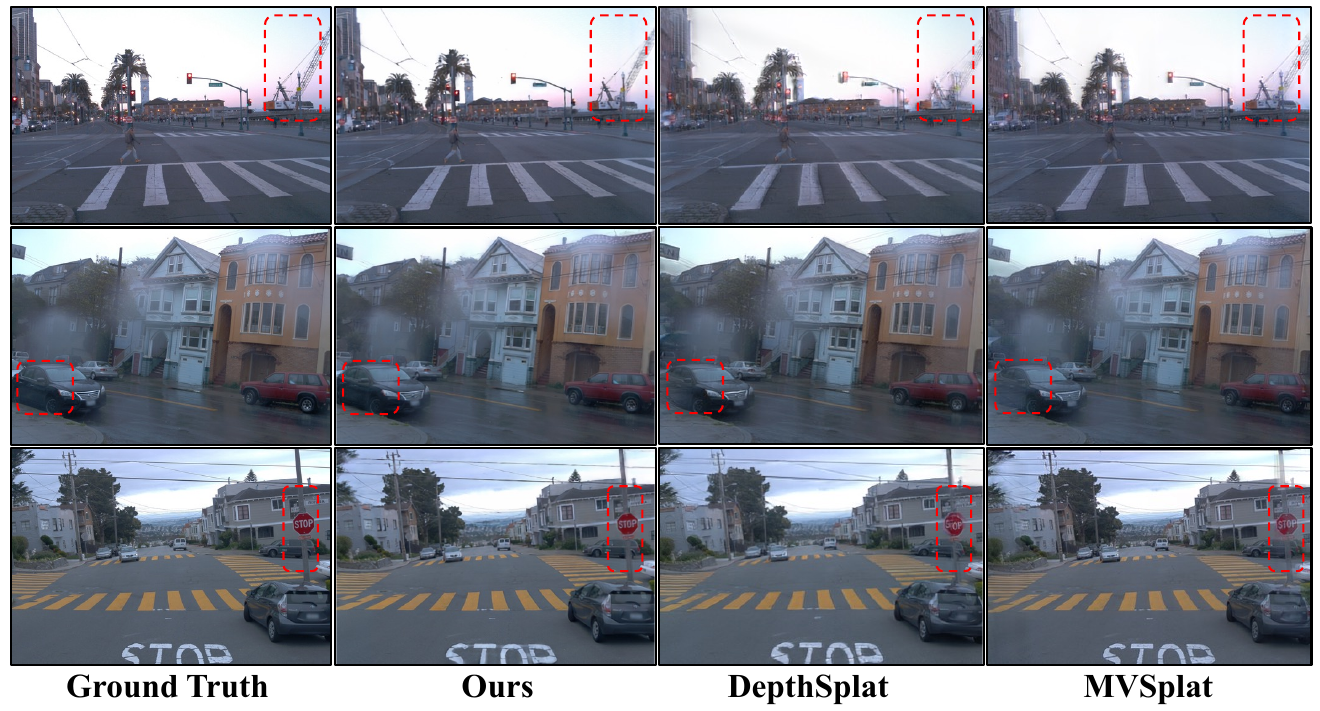}
\end{center}
\vspace{-15pt}
\caption{Qualitative comparisons on the Waymo dataset. Our method yields more detailed and consistent geometry than existing works. \textcolor{red}{Red boxes} indicate artifacts. Best viewed zoomed in.
}
\label{pic:compare}
\end{figure}

\textbf{Waymo.} We compare UniSplat against state-of-the-art sparse-view reconstruction methods, including MVSplat~\citep{chen2024mvsplat}, DepthSplat~\citep{xu2025depthsplat}, EvolSplat~\citep{miao2025evolsplat}, and DriveRecon~\citep{lu2024drivingrecon}. For the general methods MVSplat and DepthSplat, we retrain them on the Waymo Open Dataset using their official codebases. For driving-specific methods EvolSplat and DriveRecon, we conduct evaluation on our validation scenes and resize their outputs to match the resolution for fair comparison. The quantitative results are summarized in Table~\ref{tab:waymo}. UniSplat consistently outperforms all baselines across every metric for both input view reconstruction and novel view synthesis. The qualitative comparisons are shown in Figure~\ref{pic:compare}. Notably, MVSplat and DepthSplat struggle to reconstruct fine geometric details and exhibit noticeable artifacts, especially in overlapping regions between adjacent cameras. In contrast, our method produces visually coherent and high-quality results. We also report an variant (denoted by $\dagger$), in which per-camera scales are set to optimal values derived from LiDAR pointmap, leading to additional improvements.

\textbf{NuScenes.} Following \citet{wei2025omni}, we evaluate UniSplat on the nuScenes benchmark under the same protocol. As shown in Table~\ref{tab:nuc}, UniSplat surpasses the previous state of the art, Omni-Scene, achieving 25.37 PSNR (+1.10dB). 

\begin{figure}[h]
\begin{center}
\includegraphics[width=0.99\textwidth]{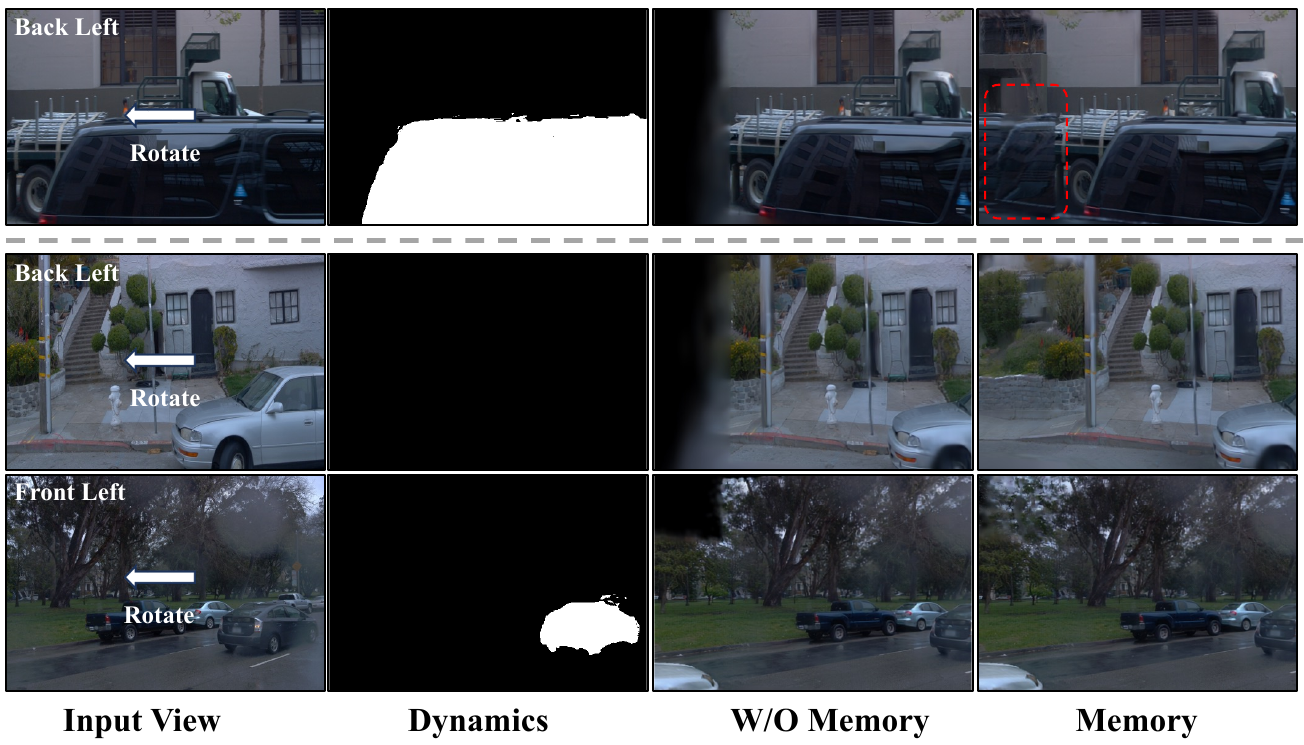}
\end{center}
\vspace{-15pt}
\caption{Qualitative results of scene completion on the Waymo dataset. \textbf{Top}: Aggregated scene without dynamic filtering, where \textcolor{red}{red boxes} indicate ghosting artifacts caused by accumulating the dynamic car. \textbf{Bottom}: Our method, equipped with dynamic-aware Gaussians, completes unobserved regions due to limited sensor coverage and bridges cross-camera gaps while avoiding dynamic artifacts. The predicted dynamic masks used for filtering are shown for reference.}
\vspace{-10pt}
\label{pic:dyn-vis}
\end{figure}

\textbf{Dynamic-aware Gaussian Completion.} UniSplat predicts per-Gaussian dynamic attributes, enabling the progressive construction of the scene during inference without manual labels. As shown in Figure~\ref{pic:dyn-vis}, the top section presents a baseline without dynamic filtering, where ghosting artifacts arise from accumulated dynamic objects. In contrast, our approach effectively completes missing regions while suppressing such artifacts. As illustrated in the bottom section, UniSplat successfully completes unobserved areas arising from two typical cases: limited 360° coverage in Waymo's five-camera setup and cross-camera blind spots. Moreover, we can observe our model clearly separates dynamic vehicles from parked ones, demonstrating its effective use of temporal context.

\subsection{Ablation Study}
In this section, we conduct ablation studies on the Waymo Open Dataset~\citep{sun2020scalability} to investigate the individual components of our framework, with a focus on novel view synthesis performance. For efficiency, we subsample the first 20\% of frames from each sequence and apply optimal scale alignment to the point map to accelerate model convergence. All models are trained for 20 epochs with a batch size of 32 on 16 GPUs. 

\textbf{Ablation on Geometric and Semantic Features in Scaffold.} Table~\ref{tab:ablate-geo-sem} investigates the contribution of geometric and semantic features from foundation models to the scaffold representation. The absence of semantic features causes a severe decline in LPIPS, increasing the error by 0.05, which can be attributed to the fact that LPIPS measures perceptual similarity using deep semantic representations. In contrast, the $2^{nd}$ and $3^{rd}$ rows show that performance gap is less pronounced when only DINO features are used, suggesting that current large-scale pretrained 2D foundation model~\citep{simeoni2025dinov3} may implicitly capture certain geometric priors.

\textbf{Analysis of Spatio-Temporal Fusion.} We ablate the effects of our spatial and temporal scaffold fusion, with results summarized in Table~\ref{tab:ablate-spatio-temporal}.  As shown in $1^{st}$ and $2^{nd}$ rows, the incorporation of spatial scaffold fusion, which aggregates spatial information in 3D space, improves performance by +0.36dB in PSNR and +0.02 in SSIM compared to the baseline that only relies on image-domain fusion. Further integration of temporal scaffold fusion, which incorporates historical context through ego-motion warping and fusion in the latent scaffold domain, brings an additional gain of +0.58dB in PSNR and +0.04 in SSIM. We also compare against a variant that explicitly uses two consecutive frames without latent-space temporal propagation. This approach achieves a lower PSNR of 24.72dB, likely due to its limited ability to model dynamic elements and restricted temporal context. These results demonstrate the effectiveness of our unified spatio-temporal modeling approach that operates directly within the 3D scaffold representation for handling sparse, minimally-overlapping camera views and complex dynamic driving scenes.

\textbf{Dual-Branch Gaussian Decoder.} We validate our dual-branch decoder design in Table~\ref{tab:ablate-two-branch}. Using only point-anchored Gaussians results in a performance degradation of 0.46 in PSNR, 0.02 in SSIM, and an increase of 0.08 in LPIPS error, underscoring the critical role of voxel-generated Gaussians in improving scene completeness by effectively filling sparsely covered regions. The voxel-only variant is excluded from comparison as it fails catastrophically at long-range rendering~\citep{wei2025omni},  yielding consistently poor performance across all metrics.

\textbf{Geometry Foundation Model.} In Table \ref{tab:ablate-geo-foundation}, We ablate the impact of the geometry foundation model on our framework's performance. Specifically, replacing the default model with MoGe-2~\citep{wang2025moge}, a recently introduced open-domain geometry estimation method, yields consistent performance, which indicates that our approach is robust to the choice of the underlying geometry foundation model. 
Notably, we exclude the representative VGGT~\citep{wang2025vggt}, as our empirical observations indicate that it generalizes less effectively than $\pi^3$ in outdoor driving scenarios.

\begin{table}
\centering
\small
\begin{minipage}{0.49\textwidth}
\centering
\caption{Impact of feature composition of $\mathbf{F}_t$. ``Geo'' and ``Sem'' denote geometric and semantic features, respectively.}
\label{tab:ablate-geo-sem}
\setlength{\tabcolsep}{6pt} 
\begin{tabular}{cc|ccc}
\hline
Geo & Sem & PSNR$\uparrow$ & SSIM$\uparrow$ & LPIPS$\downarrow$  \\
\hline
\checkmark &  & 24.78 & 0.73 & 0.35   \\
 & \checkmark & 24.85& 0.72& 0.31 \\
\checkmark & \checkmark &\textbf{25.08} & \textbf{0.74}& \textbf{0.30} \\
\hline
\end{tabular}
\end{minipage}
\hfill
\begin{minipage}{0.49\textwidth}
\centering
\setlength{\tabcolsep}{6pt}
\caption{Analysis of spatio-temporal fusion. ``Spa'' and ``Tem'' denote spatial and temporal fusion, respectively.}
\label{tab:ablate-spatio-temporal}
\begin{tabular}{cc|ccc}
\hline
Spa & Tem & PSNR$\uparrow$ & SSIM$\uparrow$ & LPIPS$\downarrow$  \\
\hline
&  & 24.14 & 0.68 &   0.32 \\
 \checkmark &  & 24.50 & 0.70 & 0.32 \\
 \checkmark &  \checkmark &\textbf{25.08} & \textbf{0.74}& \textbf{0.30} \\
\hline
\end{tabular}
\end{minipage}
\vspace{-15pt}
\end{table}

\begin{table}
\centering
\small
\begin{minipage}{0.52\textwidth}
\centering
\caption{Ablation study on the two branches of our Gaussian decoder.}
\label{tab:ablate-two-branch}
\setlength{\tabcolsep}{3.5pt} 
\begin{tabular}{cc|ccc}
\hline
Point & Voxel & PSNR$\uparrow$ & SSIM$\uparrow$ & LPIPS$\downarrow$  \\
\hline
\checkmark &  & 24.62 & 0.72 & 0.38   \\
\checkmark & \checkmark &\textbf{25.08} & \textbf{0.74}& \textbf{0.30} \\
\hline
\end{tabular}
\end{minipage}
\hfill
\begin{minipage}{0.46\textwidth}
\centering
\caption{Performance comparison of different geometry foundation models.}
\label{tab:ablate-geo-foundation}
\setlength{\tabcolsep}{6pt} 
\begin{tabular}{c|ccc}
\hline
Models & PSNR$\uparrow$ & SSIM$\uparrow$ & LPIPS$\downarrow$  \\
\hline
MoGe-2 & 24.98 & 0.74 & \textbf{0.29} \\
$\pi^3$  &\textbf{25.08} & \textbf{0.74}& 0.30 \\
\hline
\end{tabular}
\end{minipage}
\vspace{-10pt}
\end{table}


\section{Conclusion}
We presented \textbf{UniSplat}, a unified feed-forward framework for dynamic driving scene reconstruction and novel view synthesis. Our core contribution is the introduction of a 3D latent scaffold that seamlessly unifies spatio-temporal fusion from multi-camera videos. By leveraging foundation models, this scaffold encodes robust geometric and semantic priors, enabling efficient fusion directly in 3D space. We further proposed a dual-branch Gaussian decoder that generates dynamic-aware primitives from the scaffold, coupled with a streaming memory mechanism to accumulate static scene content over time for long-term completion. Extensive experiments on Waymo and nuScenes demonstrate that UniSplat not only achieves state-of-the-art performance under standard settings but also exhibits remarkable generalization to challenging viewpoints outside the original camera coverage. We believe that our framework provides a promising foundation for future research on dynamic scene understanding, interactive 4D content creation, and lifelong world modeling.

\bibliography{iclr2026_conference}

@inproceedings{Cao2025CORL, 
	author = {Wei Cao and Marcel Hallgarten and Tianyu Li and Daniel Dauner and Xunjiang Gu and Caojun Wang and Yakov Miron and Marco Aiello and Hongyang Li and Igor Gilitschenski and Boris Ivanovic and Marco Pavone and Andreas Geiger and Kashyap Chitta}, 
	title = {Pseudo-Simulation for Autonomous Driving}, 
	booktitle = {Conference on Robot Learning (CoRL)}, 
	year = {2025}, 
}

@inproceedings{yang2023unisim,
  title={Unisim: A neural closed-loop sensor simulator},
  author={Yang, Ze and Chen, Yun and Wang, Jingkang and Manivasagam, Sivabalan and Ma, Wei-Chiu and Yang, Anqi Joyce and Urtasun, Raquel},
  booktitle={Proceedings of the IEEE/CVF Conference on Computer Vision and Pattern Recognition},
  year={2023}
}

@inproceedings{tonderski2024neurad,
  title={Neurad: Neural rendering for autonomous driving},
  author={Tonderski, Adam and Lindstr{\"o}m, Carl and Hess, Georg and Ljungbergh, William and Svensson, Lennart and Petersson, Christoffer},
  booktitle={Proceedings of the IEEE/CVF Conference on Computer Vision and Pattern Recognition},
  year={2024}
}

@inproceedings{huang2024gaussianformer,
  title={Gaussianformer: Scene as gaussians for vision-based 3d semantic occupancy prediction},
  author={Huang, Yuanhui and Zheng, Wenzhao and Zhang, Yunpeng and Zhou, Jie and Lu, Jiwen},
  booktitle={European Conference on Computer Vision},
  year={2024},
}

@inproceedings{huang2025gaussianformer,
  title={Gaussianformer-2: Probabilistic gaussian superposition for efficient 3d occupancy prediction},
  author={Huang, Yuanhui and Thammatadatrakoon, Amonnut and Zheng, Wenzhao and Zhang, Yunpeng and Du, Dalong and Lu, Jiwen},
  booktitle={Proceedings of the Computer Vision and Pattern Recognition Conference},
  year={2025}
}

@inproceedings{murai2025mast3r,
  title={MASt3R-SLAM: Real-time dense SLAM with 3D reconstruction priors},
  author={Murai, Riku and Dexheimer, Eric and Davison, Andrew J},
  booktitle={Proceedings of the Computer Vision and Pattern Recognition Conference},
  year={2025}
}

@inproceedings{yan2024street,
  title={Street gaussians: Modeling dynamic urban scenes with gaussian splatting},
  author={Yan, Yunzhi and Lin, Haotong and Zhou, Chenxu and Wang, Weijie and Sun, Haiyang and Zhan, Kun and Lang, Xianpeng and Zhou, Xiaowei and Peng, Sida},
  booktitle={European Conference on Computer Vision},
  year={2024},
}

@article{xu2025adgs,
    title={{AD-GS}: Object-Aware {B-Spline} {Gaussian} Splatting for Self-Supervised Autonomous Driving},
    author={Jiawei, Xu and Kai, Deng and Zexin, Fan and Shenlong, Wang and Jin, Xie and Jian, Yang},
    journal={International Conference on Computer Vision},
    year={2025},
}

@article{kerbl20233d,
  title={3D Gaussian splatting for real-time radiance field rendering},
  author={Kerbl, Bernhard and Kopanas, Georgios and Leimk{\"u}hler, Thomas and Drettakis, George},
  journal={ACM Trans. Graph.},
  year={2023}
}

@inproceedings{xu2025depthsplat,
  title={Depthsplat: Connecting gaussian splatting and depth},
  author={Xu, Haofei and Peng, Songyou and Wang, Fangjinhua and Blum, Hermann and Barath, Daniel and Geiger, Andreas and Pollefeys, Marc},
  booktitle={Proceedings of the Computer Vision and Pattern Recognition Conference},
  year={2025}
}

@inproceedings{chen2024mvsplat,
  title={Mvsplat: Efficient 3d gaussian splatting from sparse multi-view images},
  author={Chen, Yuedong and Xu, Haofei and Zheng, Chuanxia and Zhuang, Bohan and Pollefeys, Marc and Geiger, Andreas and Cham, Tat-Jen and Cai, Jianfei},
  booktitle={European Conference on Computer Vision},
  year={2024},
  organization={Springer}
}

@article{lu2024drivingrecon,
  title={DrivingRecon: Large 4D Gaussian Reconstruction Model For Autonomous Driving},
  author={Lu, Hao and Xu, Tianshuo and Zheng, Wenzhao and Zhang, Yunpeng and Zhan, Wei and Du, Dalong and Tomizuka, Masayoshi and Keutzer, Kurt and Chen, Yingcong},
  journal={arXiv preprint arXiv:2412.09043},
  year={2024}
}

@inproceedings{miao2025evolsplat,
  title={Evolsplat: Efficient volume-based gaussian splatting for urban view synthesis},
  author={Miao, Sheng and Huang, Jiaxin and Bai, Dongfeng and Yan, Xu and Zhou, Hongyu and Wang, Yue and Liu, Bingbing and Geiger, Andreas and Liao, Yiyi},
  booktitle={Proceedings of the Computer Vision and Pattern Recognition Conference},
  year={2025}
}

@inproceedings{wei2025omni,
  title={Omni-scene: Omni-gaussian representation for ego-centric sparse-view scene reconstruction},
  author={Wei, Dongxu and Li, Zhiqi and Liu, Peidong},
  booktitle={Proceedings of the Computer Vision and Pattern Recognition Conference},
  year={2025}
}

@inproceedings{wang2025vggt,
  title={Vggt: Visual geometry grounded transformer},
  author={Wang, Jianyuan and Chen, Minghao and Karaev, Nikita and Vedaldi, Andrea and Rupprecht, Christian and Novotny, David},
  booktitle={Proceedings of the Computer Vision and Pattern Recognition Conference},
  year={2025}
}

@article{wang2025pi,
  title={$\backslash\pi^3$: Scalable Permutation-Equivariant Visual Geometry Learning},
  author={Wang, Yifan and Zhou, Jianjun and Zhu, Haoyi and Chang, Wenzheng and Zhou, Yang and Li, Zizun and Chen, Junyi and Pang, Jiangmiao and Shen, Chunhua and He, Tong},
  journal={arXiv preprint arXiv:2507.13347},
  year={2025}
}

@inproceedings{sun2020scalability,
  title={Scalability in perception for autonomous driving: Waymo open dataset},
  author={Sun, Pei and Kretzschmar, Henrik and Dotiwalla, Xerxes and Chouard, Aurelien and Patnaik, Vijaysai and Tsui, Paul and Guo, James and Zhou, Yin and Chai, Yuning and Caine, Benjamin and others},
  booktitle={Proceedings of the IEEE/CVF conference on computer vision and pattern recognition},
  year={2020}
}

@inproceedings{caesar2020nuscenes,
  title={nuscenes: A multimodal dataset for autonomous driving},
  author={Caesar, Holger and Bankiti, Varun and Lang, Alex H and Vora, Sourabh and Liong, Venice Erin and Xu, Qiang and Krishnan, Anush and Pan, Yu and Baldan, Giancarlo and Beijbom, Oscar},
  booktitle={Proceedings of the IEEE/CVF conference on computer vision and pattern recognition},
  year={2020}
}

@article{mildenhall2021nerf,
  title={Nerf: Representing scenes as neural radiance fields for view synthesis},
  author={Mildenhall, Ben and Srinivasan, Pratul P and Tancik, Matthew and Barron, Jonathan T and Ramamoorthi, Ravi and Ng, Ren},
  journal={Communications of the ACM},
  year={2021},
}

@inproceedings{yu2024mip,
  title={Mip-splatting: Alias-free 3d gaussian splatting},
  author={Yu, Zehao and Chen, Anpei and Huang, Binbin and Sattler, Torsten and Geiger, Andreas},
  booktitle={Proceedings of the IEEE/CVF conference on computer vision and pattern recognition},
  year={2024}
}

@inproceedings{hu2023tri,
  title={Tri-miprf: Tri-mip representation for efficient anti-aliasing neural radiance fields},
  author={Hu, Wenbo and Wang, Yuling and Ma, Lin and Yang, Bangbang and Gao, Lin and Liu, Xiao and Ma, Yuewen},
  booktitle={Proceedings of the IEEE/CVF International Conference on Computer Vision},
  year={2023}
}

@inproceedings{yang2025improving,
  title={Improving Gaussian Splatting with Localized Points Management},
  author={Yang, Haosen and Zhang, Chenhao and Wang, Wenqing and Volino, Marco and Hilton, Adrian and Zhang, Li and Zhu, Xiatian},
  booktitle={Proceedings of the Computer Vision and Pattern Recognition Conference},
  year={2025}
}

@article{muller2022instant,
  title={Instant neural graphics primitives with a multiresolution hash encoding},
  author={M{\"u}ller, Thomas and Evans, Alex and Schied, Christoph and Keller, Alexander},
  journal={ACM transactions on graphics (TOG)},
  year={2022},
}

@inproceedings{xu2024murf,
  title={MuRF: multi-baseline radiance fields},
  author={Xu, Haofei and Chen, Anpei and Chen, Yuedong and Sakaridis, Christos and Zhang, Yulun and Pollefeys, Marc and Geiger, Andreas and Yu, Fisher},
  booktitle={Proceedings of the IEEE/CVF Conference on Computer Vision and Pattern Recognition},
  year={2024}
}

@inproceedings{charatan2024pixelsplat,
  title={pixelsplat: 3d gaussian splats from image pairs for scalable generalizable 3d reconstruction},
  author={Charatan, David and Li, Sizhe Lester and Tagliasacchi, Andrea and Sitzmann, Vincent},
  booktitle={Proceedings of the IEEE/CVF conference on computer vision and pattern recognition},
  year={2024}
}

@inproceedings{szymanowicz2024splatter,
  title={Splatter image: Ultra-fast single-view 3d reconstruction},
  author={Szymanowicz, Stanislaw and Rupprecht, Chrisitian and Vedaldi, Andrea},
  booktitle={Proceedings of the IEEE/CVF conference on computer vision and pattern recognition},
  year={2024}
}

@inproceedings{yan2025streetcrafter,
  title={Streetcrafter: Street view synthesis with controllable video diffusion models},
  author={Yan, Yunzhi and Xu, Zhen and Lin, Haotong and Jin, Haian and Guo, Haoyu and Wang, Yida and Zhan, Kun and Lang, Xianpeng and Bao, Hujun and Zhou, Xiaowei and others},
  booktitle={Proceedings of the Computer Vision and Pattern Recognition Conference},
  year={2025}
}

@inproceedings{fan2025freesim,
  title={Freesim: Toward free-viewpoint camera simulation in driving scenes},
  author={Fan, Lue and Zhang, Hao and Wang, Qitai and Li, Hongsheng and Zhang, Zhaoxiang},
  booktitle={Proceedings of the Computer Vision and Pattern Recognition Conference},
  year={2025}
}

@inproceedings{tian2025drivingforward,
  title={Drivingforward: Feed-forward 3d gaussian splatting for driving scene reconstruction from flexible surround-view input},
  author={Tian, Qijian and Tan, Xin and Xie, Yuan and Ma, Lizhuang},
  booktitle={Proceedings of the AAAI Conference on Artificial Intelligence},
  year={2025}
}

@article{chen2023periodic,
  title={Periodic vibration gaussian: Dynamic urban scene reconstruction and real-time rendering},
  author={Chen, Yurui and Gu, Chun and Jiang, Junzhe and Zhu, Xiatian and Zhang, Li},
  journal={arXiv preprint arXiv:2311.18561},
  year={2023}
}

@article{ren2024scube,
  title={Scube: Instant large-scale scene reconstruction using voxsplats},
  author={Ren, Xuanchi and Lu, Yifan and Liang, Hanxue and Wu, Zhangjie and Ling, Huan and Chen, Mike and Fidler, Sanja and Williams, Francis and Huang, Jiahui},
  journal={Advances in Neural Information Processing Systems},
  year={2024}
}

@inproceedings{wang2024dust3r,
  title={Dust3r: Geometric 3d vision made easy},
  author={Wang, Shuzhe and Leroy, Vincent and Cabon, Yohann and Chidlovskii, Boris and Revaud, Jerome},
  booktitle={Proceedings of the IEEE/CVF Conference on Computer Vision and Pattern Recognition},
  year={2024}
}

@inproceedings{yang2025fast3r,
  title={Fast3r: Towards 3d reconstruction of 1000+ images in one forward pass},
  author={Yang, Jianing and Sax, Alexander and Liang, Kevin J and Henaff, Mikael and Tang, Hao and Cao, Ang and Chai, Joyce and Meier, Franziska and Feiszli, Matt},
  booktitle={Proceedings of the Computer Vision and Pattern Recognition Conference},
  year={2025}
}

@inproceedings{wang2025continuous,
  title={Continuous 3d perception model with persistent state},
  author={Wang, Qianqian and Zhang, Yifei and Holynski, Aleksander and Efros, Alexei A and Kanazawa, Angjoo},
  booktitle={Proceedings of the Computer Vision and Pattern Recognition Conference},
  year={2025}
}

@article{chen2025long3r,
  title={LONG3R: Long Sequence Streaming 3D Reconstruction},
  author={Chen, Zhuoguang and Qin, Minghui and Yuan, Tianyuan and Liu, Zhe and Zhao, Hang},
  journal={arXiv preprint arXiv:2507.18255},
  year={2025}
}

@article{huang2024s3gaussian,
        title={S3Gaussian: Self-Supervised Street Gaussians for Autonomous Driving},
        author={Huang, Nan and Wei, Xiaobao and Zheng, Wenzhao and An, Pengju and Lu, Ming and Zhan, Wei and Tomizuka,    Masayoshi and Keutzer, Kurt and Zhang, Shanghang},
        journal={arXiv preprint arXiv:2405.20323},
        year={2024}
      }

@inproceedings{zhao2025drivedreamer4d,
  title={Drivedreamer4d: World models are effective data machines for 4d driving scene representation},
  author={Zhao, Guosheng and Ni, Chaojun and Wang, Xiaofeng and Zhu, Zheng and Zhang, Xueyang and Wang, Yida and Huang, Guan and Chen, Xinze and Wang, Boyuan and Zhang, Youyi and others},
  booktitle={Proceedings of the Computer Vision and Pattern Recognition Conference},
  year={2025}
}

@inproceedings{zhou2024feature,
  title={Feature 3dgs: Supercharging 3d gaussian splatting to enable distilled feature fields},
  author={Zhou, Shijie and Chang, Haoran and Jiang, Sicheng and Fan, Zhiwen and Zhu, Zehao and Xu, Dejia and Chari, Pradyumna and You, Suya and Wang, Zhangyang and Kadambi, Achuta},
  booktitle={Proceedings of the IEEE/CVF Conference on Computer Vision and Pattern Recognition},
  year={2024}
}

@article{zuo2025fmgs,
  title={Fmgs: Foundation model embedded 3d gaussian splatting for holistic 3d scene understanding},
  author={Zuo, Xingxing and Samangouei, Pouya and Zhou, Yunwen and Di, Yan and Li, Mingyang},
  journal={International Journal of Computer Vision},
  year={2025},
  publisher={Springer}
}

@article{wang2025moge,
  title={MoGe-2: Accurate Monocular Geometry with Metric Scale and Sharp Details},
  author={Wang, Ruicheng and Xu, Sicheng and Dong, Yue and Deng, Yu and Xiang, Jianfeng and Lv, Zelong and Sun, Guangzhong and Tong, Xin and Yang, Jiaolong},
  journal={arXiv preprint arXiv:2507.02546},
  year={2025}
}

@inproceedings{wang2025mogev1,
  title={Moge: Unlocking accurate monocular geometry estimation for open-domain images with optimal training supervision},
  author={Wang, Ruicheng and Xu, Sicheng and Dai, Cassie and Xiang, Jianfeng and Deng, Yu and Tong, Xin and Yang, Jiaolong},
  booktitle={Proceedings of the Computer Vision and Pattern Recognition Conference},
  pages={5261--5271},
  year={2025}
}

@article{oquab2023dinov2,
  title={Dinov2: Learning robust visual features without supervision},
  author={Oquab, Maxime and Darcet, Timoth{\'e}e and Moutakanni, Th{\'e}o and Vo, Huy and Szafraniec, Marc and Khalidov, Vasil and Fernandez, Pierre and Haziza, Daniel and Massa, Francisco and El-Nouby, Alaaeldin and others},
  journal={arXiv preprint arXiv:2304.07193},
  year={2023}
}

@inproceedings{zhang2018unreasonable,
  title={The unreasonable effectiveness of deep features as a perceptual metric},
  author={Zhang, Richard and Isola, Phillip and Efros, Alexei A and Shechtman, Eli and Wang, Oliver},
  booktitle={Proceedings of the IEEE conference on computer vision and pattern recognition},
  year={2018}
}

@article{wang2004image,
  title={Image quality assessment: from error visibility to structural similarity},
  author={Wang, Zhou and Bovik, Alan C and Sheikh, Hamid R and Simoncelli, Eero P},
  journal={IEEE transactions on image processing},
  year={2004},
}

@inproceedings{
yang2024emernerf,
title={EmerNe{RF}: Emergent Spatial-Temporal Scene Decomposition via Self-Supervision},
author={Jiawei Yang and Boris Ivanovic and Or Litany and Xinshuo Weng and Seung Wook Kim and Boyi Li and Tong Che and Danfei Xu and Sanja Fidler and Marco Pavone and Yue Wang},
booktitle={The Twelfth International Conference on Learning Representations},
year={2024},
}

@inproceedings{
loshchilov2018decoupled,
title={Decoupled Weight Decay Regularization},
author={Ilya Loshchilov and Frank Hutter},
booktitle={International Conference on Learning Representations},
year={2019},
}

@article{simeoni2025dinov3,
  title={Dinov3},
  author={Sim{\'e}oni, Oriane and Vo, Huy V and Seitzer, Maximilian and Baldassarre, Federico and Oquab, Maxime and Jose, Cijo and Khalidov, Vasil and Szafraniec, Marc and Yi, Seungeun and Ramamonjisoa, Micha{\"e}l and others},
  journal={arXiv preprint arXiv:2508.10104},
  year={2025}
}

@inproceedings{zhang2025transplat,
  title={Transplat: Generalizable 3d gaussian splatting from sparse multi-view images with transformers},
  author={Zhang, Chuanrui and Zou, Yingshuang and Li, Zhuoling and Yi, Minmin and Wang, Haoqian},
  booktitle={Proceedings of the AAAI Conference on Artificial Intelligence},
  year={2025}
}

@inproceedings{xu2022point,
  title={Point-nerf: Point-based neural radiance fields},
  author={Xu, Qiangeng and Xu, Zexiang and Philip, Julien and Bi, Sai and Shu, Zhixin and Sunkavalli, Kalyan and Neumann, Ulrich},
  booktitle={Proceedings of the IEEE/CVF conference on computer vision and pattern recognition},
  year={2022}
}

@inproceedings{zhou2024drivinggaussian,
  title={Drivinggaussian: Composite gaussian splatting for surrounding dynamic autonomous driving scenes},
  author={Zhou, Xiaoyu and Lin, Zhiwei and Shan, Xiaojun and Wang, Yongtao and Sun, Deqing and Yang, Ming-Hsuan},
  booktitle={Proceedings of the IEEE/CVF conference on computer vision and pattern recognition},
  year={2024}
}

@inproceedings{xiao2025spatialtrackerv2,
  title={Spatialtrackerv2: 3d point tracking made easy},
  author={Xiao, Yuxi and Wang, Jianyuan and Xue, Nan and Karaev, Nikita and Makarov, Yuri and Kang, Bingyi and Zhu, Xing and Bao, Hujun and Shen, Yujun and Zhou, Xiaowei},
  booktitle={Proceedings of the IEEE/CVF international conference on computer vision},
  year={2025}
}

@article{yan2025st,
  title={ST-GS: Vision-Based 3D Semantic Occupancy Prediction with Spatial-Temporal Gaussian Splatting},
  author={Yan, Xiaoyang and Pei, Muleilan and Shen, Shaojie},
  journal={arXiv preprint arXiv:2509.16552},
  year={2025}
}
\bibliographystyle{iclr2026_conference}

\end{document}